\documentclass[final]{l4dc2025}

\title[Multi-Constraint Safe Reinforcement Learning]{Multi-Constraint Safe Reinforcement Learning via Closed-form Solution for Log-Sum-Exp Approximation of Control Barrier Functions}
\usepackage{times}
\usepackage{multirow}
\usepackage{caption}

\author{%
	\Name{Chenggang Wang} \Email{cgwang-auv@sjtu.edu.cn}\\
	\addr Shanghai Jiao Tong University, Shanghai, China
	\AND
	\Name{Xinyi Wang} \Email{xinywa@umich.edu}\\
	\addr University of Michigan,  Ann Arbor, MI, USA
	\AND
	\Name{Yutong Dong} \Email{755467293@sjtu.edu.cn}\\
	\addr Shanghai Jiao Tong University, Shanghai, China
	\AND
	\Name{Lei Song} \Email{songlei\_24@sjtu.edu.cn}\\
	\addr Shanghai Jiao Tong University, Shanghai, China
	\AND
	\Name{Xinping Guan} \Email{xpguan@sjtu.edu.cn}\\
	\addr Shanghai Jiao Tong University, Shanghai, China
}

\begin{document}

\maketitle

\begin{abstract}%
 The safety of training task policies and their subsequent application using reinforcement learning (RL) methods has become a focal point in the field of safe RL. A central challenge in this area remains the establishment of theoretical guarantees for safety during both the learning and deployment processes.
 Given the successful implementation of Control Barrier Function (CBF)-based safety strategies in a range of control-affine robotic systems, CBF-based safe RL demonstrates significant promise for practical applications in real-world scenarios.
 However, integrating these two approaches presents several challenges. First, embedding safety optimization within the RL training pipeline requires that the optimization outputs be differentiable with respect to the input parameters, a condition commonly referred to as differentiable optimization, which is non-trivial to solve. 
 Second, the differentiable optimization framework confronts significant efficiency issues, especially when dealing with multi-constraint problems.
 To address these challenges, this paper presents a CBF-based safe RL architecture that effectively mitigates the issues outlined above. The proposed approach constructs a continuous AND logic approximation for the multiple constraints using a single composite CBF. By leveraging this approximation, a close-form solution of the quadratic programming is derived for the policy network in RL, thereby circumventing the need for differentiable optimization within the end-to-end safe RL pipeline. 
 This strategy significantly reduces computational complexity because of the closed-form solution while maintaining safety guarantees.
 Simulation results demonstrate that, in comparison to existing approaches relying on differentiable optimization, the proposed method significantly reduces training computational costs while ensuring provable safety throughout the training process. This advancement opens up promising potential for applications in large-scale optimization problems.
\end{abstract}


\section{Introduction}

The safety of reinforcement learning (RL) during both training and deployment phases has garnered increasing attention \cite{Lavan24L4DC,pmlr-v242-vaskov24a,pmlr-v242-buerger24a}, particularly due to the safety-critical nature of many robotic systems. A core challenge lies in ensuring provable safety throughout these phases. Traditional RL methods commonly address safety by penalizing unsafe behaviors, which inevitably leads to the exploration of unsafe actions during training and fails to guarantee the safety of the learned policy during deployment. 
Recent solutions can be divided into two categories: constrained optimization-based methods and safety filter-based methods.
For constrained optimization involving multiple safety constraints, Lagrangian-based safe RL methods \cite{xu2021crpo,pmlr-v242-yao24a} are proposed to improve training efficiency with constraint satisfaction.
Safety filter based methods typically rely on certificate functions such as Control Barrier Functions (CBFs), or Hamilton-Jacobi Reachability value functions.
However, there remains a lack of efficient and generalizable approaches to ensure safety across all phases of the RL process.

The CBF-based approach \cite{Wang17TRO,Ames19ECC,Agrawal21CDC,Wang23RAL,Xiao22TAC} theoretically ensures safety for control strategies and has been widely applied to various control-affine robotic systems, such as autonomous vehicles \cite{Wang23RAL}, bipedal robots \cite{csomay2021episodic} and etc. 
The core idea involves formulating safety constraints for the control strategy, defining a safe set through these constraints, and deriving forward invariance conditions for the safe set to impose decision-variable constraints that ensure safety. 
These constraints are then integrated into an optimization problem to generate safe strategies. 
Typically, the safety optimization is based on system models and nominal controllers derived from control theory. Building on this framework, learning-based methods can replace nominal controllers, leveraging the powerful approximation capabilities and superior task performance of learning techniques \cite{Cheng_Orosz_Murray_Burdick_2019}.

Integrating safety optimization into the pipeline of control policy learning can be framed as a decision-focused learning paradigm \cite{Shah22NIPS}. 
In this framework, the prediction phase is handled by a RL policy network, followed by downstream safety optimization to generate the final safe strategy and evaluate its performance. 
This end-to-end approach requires the safety optimization process to be differentiable, which is often challenging due to issues like solution discontinuity \cite{ferber2020mipaal} and gradient approximation \cite{wilder2019melding}.
Recent works in decision-focused learning address these challenges through various methods: using surrogates to replace the original optimization problem and learning loss functions \cite{wilder2019melding} or constructing differentiable optimization tools \cite{amos2017optnet,NEURIPS2022_18596929,agrawal2019differentiable}. 
For control-affine systems, safety behavior optimization benefits from linear relaxation of decision variables via Nagumo's theorem \cite{Ames19ECC}, which avoids the complexity of differentiable nonlinear programming \cite{NEURIPS2022_18596929} or mixed-integer programming \cite{ferber2020mipaal}. This allows the use of differentiable Quadratic Programming (QP) solvers \cite{amos2017optnet,agrawal2019differentiable}.

Recent research on differentiable QP-based safe control \cite{Emam22RAL,Ma22ECC,amos2018differentiable,romero2024actor} primarily focuses on three aspects: (1) addressing the impact of constraint parameters, such as environmental changes on safety strategies, e.g., \cite{Ma22ECC}, by adjusting the class-$\mathcal{K}$ functions within safety constraints via differentiable QP; 
(2) constructing linear MPC problems \cite{amos2018differentiable} and tuning receding horizon parameters during optimization through differentiable QP to enhance task performance \cite{romero2024actor}; and (3) imitating safe behaviors \cite{Xiao23TRO} or integrating safe QP as the final layer of RL policy networks \cite{Emam22RAL} to generate safe optimization strategies.
Differentiable QP frameworks offer several notable advantages. First, they enable a decoupled design of task policy learning and safety correction, thereby facilitating the seamless integration of various learning methodologies. Second, the end-to-end learning of optimized strategies often yields superior task performance compared to hierarchical learning frameworks that incorporate safety corrections post-policy training. Despite these benefits, differentiable QP frameworks are not without limitations. Differentiable optimization is inherently complex, particularly for problems involving discrete decision variables. Furthermore, each gradient update requires solving an optimization problem and subsequently differentiating through it, which can result in significant computational cost.

Given these observations, this work focuses on safe RL with CBF-based optimization and addresses the computational complexity associated with differentiable optimization.
Given that safety-critical applications often involve multiple constraints within the optimization problem, a continuous Log-Sum-Exp approximation is employed to transform multiple constraints into a single composite constraint. 
Utilizing this composite constraint, the closed-form solution of the corresponding QP is derived and integrated into the final layer of the RL policy network, which enables an end-to-end training pipeline with analytical computation, effectively serving as a surrogate for the differentiable QP.
The proposed framework significantly reduces the computational cost associated with computing the derivatives of the differentiable QP output with respect to its input parameters \cite{amos2017optnet,agrawal2019differentiable}, offering an efficient and scalable solution for training in large-scale optimization problems.

\section{Preliminaries}
\subsection{Safe policy via CBF-based QP}
Consider a control-affine system
\begin{equation}\label{affine}
	\dot{x} = f(x)+g(x)u,
\end{equation}
where $x\in \mathbb{R}^n $ denotes system state, $u\in \mathbb{R}^m$ denotes control input (policy).
In this paper, we consider 
$ f : \mathbb{R}^n \rightarrow \mathbb{R}^n, g : \mathbb{R}^n \rightarrow \mathbb{R}^{n \times m}  $ are bounded Lipschitz continuous vector fields and $f,g$ are known for safety guarantee.
This consideration is common, since most of the mechanical systems can be formulated as the control-affine form including manipulators, autonomous vehicles, drones, bipedal robots, and etc.
For the safety of control-affine systems at the dynamical level, CBFs have successful applications.
The safety is related to the desired safe sets which can be defined by continuous differentiable functions $h_i(x) : \mathbb{R}^n \to \mathbb{R}, i = 1, \dots, I$:
\begin{equation}\label{Ci}
	\mathcal{C}_i \triangleq \left\lbrace x\in \mathbb{R}^n:h_i(x)\ge 0\right\rbrace ,
\end{equation}
\begin{equation}
	\partial \mathcal{C}_i \triangleq \left\lbrace x\in \mathbb{R}^n:h_i(x)=  0\right\rbrace ,
\end{equation}
\begin{equation}
	\mathrm{Int} \mathcal{C}_i \triangleq \left\lbrace x\in \mathbb{R}^n:h_i(x) > 0\right\rbrace .
\end{equation}
The set $\mathcal{C}_i$ is forward invariant if for any initial state $x(0) \in \mathcal{C}_i, x(t)\in \mathcal{C}_i, \forall t\in[0,\infty)$.
The system is safe if all $\mathcal{C}_i,i = 1, \dots, I $ are forward invariant.

Given the dynamics in \eqref{affine} and safety requirement, the forward invariance condition based on CBF
is formulated as follows:
Let $\mathcal{C}_i$ be the 0-superlevel set of a continuously differentiable function $h_i:\mathbb{R}^n \to \mathbb{R}$.
The function $h_i$ is a CBF for \eqref{affine} w.r.t. $\mathcal{C}_i$ if there exists extended class $\mathcal{K}$ function $\alpha$ and $u\in\mathbb{R}^m$ such that 
\begin{equation}
	L_fh_i(x)+L_gh_i(x)u\ge -\alpha(h_i(x)),
\end{equation}
where $L_fh_i(x)=\frac{\partial h_i(x)}{\partial x} f(x), L_gh_i(x)=\frac{\partial h_i(x)}{\partial x} g(x)$ w.r.t $f,g$.

Since all the $I$th safe constraints are linear on $u$, the control optimization based on QP can be formulated as
\begin{equation}\label{QP1}
	\begin{aligned}
		u_s &= \arg\min_{u\in\mathbb{R}^m} \frac{1}{2}\| u-\bar{u}\|_2^2 \\
		& \mathrm{s.t.}\  L_fh_i(x) +L_g h_i(x)u \ge -\alpha(h_i(x)) , i = 1, \dots, I,
	\end{aligned}
\end{equation}
where $\bar{u}$ denotes the nominal controller designed for the original task objective.
The optimization \eqref{QP1} minimally corrects the nominal controller when  $\bar{u}$ violates the safety constraints,
resulting in the safe policy $u_s$. 
More details are referred to Lemma 2 and 3 in \cite{breeden2023compositions} or Theorem 1 in \cite{aali2022multiple}, with an assumption for nonempty feasible set of $u_s$.

\subsection{Soft Actor-Critic}

As an off-policy RL algorithm, the Soft Actor-Critic (SAC) \cite{SAC} leverages its high sample efficiency and entropy regularization features to offer performance advantages in RL methods for continuous action spaces.
The entropy objective is optimized by 
\begin{equation}\label{entro}
	\begin{aligned}
		\pi^* &= \arg\max \sum_t \mathbb{E}_{(x_t,u_t^{\phi}) \thicksim \rho_{\pi}} \left[ r(x_t,u_t^{\phi}) + \alpha_e H(\pi (\cdot |x_t))               \right],
	\end{aligned}
\end{equation}

The SAC algorithm utilizes an AC approach, where the critic is represented by a Q-function parameterized by  $\theta$, and the actor is represented by a policy  $\pi$ parameterized by $\phi$. The critic loss $J_Q(\theta)$ aims to minimize the difference between the Q-values generated by the critic and the sum of the rewards plus the expected value of the next state's value function:
\begin{equation}
	\begin{aligned}
		J_Q(\theta) = \mathbb{E}_{(x_t, u_t^{\phi}) \sim D_r} \left[ \frac{1}{2} \left( Q_\theta (x_t, u_t) - \left( r(x_t, u^{\phi}_t) \right. \right. \right. \\ 
		\left. \left. \left. + \gamma \mathbb{E}_{x_{t+1} \sim p} \left[ V_{\hat{\theta}}(x_{t+1}) \right] \right) \right)^2 \right],
	\end{aligned}
\end{equation}
where $D_r$ is the replay buffer, and $\hat{\theta}$ represents the target Q-network parameters. 
The replay buffer $D_r$ provides a diverse set of experiences, enabling the critic to learn from a broad range of past states and actions, which enhances sample efficiency. The target Q-network parameters $\hat{\theta}$ ensure stable updates by serving as a slowly updating reference.

The entropy term is included to promote exploration and prevent premature convergence to suboptimal policies, which is given by:
\begin{equation}
	H(\pi (\cdot |x_t)) =- \log \pi_\phi \left( u_t^{\phi} | x_t \right).
\end{equation}

The policy loss encourages actions that maximize both the expected reward and the entropy, leading to an effective balance between performance and exploration, which is given by
\begin{equation}\label{Policy_loss}
	\begin{aligned}
		J_{\pi}(\phi) &=  \mathbb{E}_{x_t\thicksim \mathcal{D}_r} \left[ \mathbb{E}_{u_t^{\phi}\thicksim \pi_{\phi}}  [\alpha_e\log \pi_{\phi}(u_t^{\phi}|x_t)-Q_{\theta}(x_t,u_t^{\phi})]               \right].
	\end{aligned}
\end{equation}

One of the primary advantages of constructing a differentiable optimization framework is the ability to decouple the design of the safety layer, enabling seamless integration into the policy network of actor-critic (AC)-based RL methods.
Therefore, the SAC serves as a candidate when implementing a differentiable QP layer. 
The policy loss is given by
\begin{equation}\label{Policy_loss_QP}
	\begin{aligned}
		J_{\pi}(\phi) &=  \mathbb{E}_{x_t\thicksim \mathcal{D}_r} \\ 
		&\left[ \mathbb{E}_{u_t^{\phi}\thicksim \pi_{\phi}}  [\alpha_e\log \pi_{\phi}(u_t^{\phi}|x_t)-Q_{\theta}(x_t,u_t^{\phi}+u_t^C)]               \right],
	\end{aligned}
\end{equation}
where $u_t^C$ is the compensation term computed by differentiable QP layer.

\section{Main results}
\subsection{Composite CBF for multiple constraints}

To solve the CBF-based optimization under multiple constraints, the constraints are regarded as the intersection of safe sets defined by these CBFs. 
Each safe constraint $h_i$ is defined by a 0-superlevel set and their intersection is defined as
\begin{equation}\label{intersection}
	\bigcap_{i = 1, \dots, I} C_i = \{ x \in \mathbb{R}^n : h_i(x) \geq 0  \},
\end{equation}
where $I$ denotes the number of the safety constraints.

In other words, the intersection of sets captures the logical AND relationship between multiple safety constraints, which is denoted as
\begin{equation}
	x \in \bigcap_{i = 1, \dots, I} C_i \iff x \in C_1 \text{ AND } x \in C_2 \cdots \text{ AND } x \in C_I.
\end{equation}

When there are multiple constraints, the complexity of the QP problem increases, generally making it impossible to derive a closed-form solution, thus requiring numerical optimization methods such as active set or interior point methods. 
However, inspired by existing literature \cite{Molnar23CSL} solving complex safety specifications, this paper employs a Log-Sum-Exp approximation technique to transform multiple constraints into a single constraint, thereby enabling a closed-form solution for the safe QP.

The approximated composite single CBF is constructed as:
\begin{equation}\label{composite}
	h(x) = -\frac{1}{\kappa} \ln \left( \sum_{i=1}^I e^{-\kappa h_i(x)} \right),
\end{equation}
whose Lie derivatives are expressed by:
\begin{equation}
	L_f h(x) = \sum_{i=1}^I \lambda_i(x) L_f h_i(x), \quad L_g h(x) = \sum_{i=1}^I \lambda_i(x) L_g h_i(x),
\end{equation}
where
\begin{equation}
	\lambda_i(x) = e^{-\kappa (h_i(x) - h(x))},
\end{equation}
with  $\sum_{i \in I} \lambda_i(x) = 1$ and $\kappa > 0$. 

Since an equivalent substitution for the constraints of optimization problem \eqref{QP1} is $\min_{}h_i(x)\ge 0, i=1,\cdots,I$.
The composite CBF in \eqref{composite} shares the following property.

\textbf{Lemma 1:} \cite{Molnar23CSL}
Consider sets $C_i$ in \eqref{Ci} and their intersection in \eqref{intersection}. 
Continuous function $h(x)$ in \eqref{composite} under approximates $\min_{i=1,\cdots,I}h_i(x)\ge 0$ with bounds:
\begin{equation}
	\min_{i=1,\cdots,I} h_i(x) - \frac{\ln I}{\kappa} \leq h(x) \leq \min_{i=1,\cdots,I} h_i(x) \quad \forall x \in \mathbb{R}^n,
\end{equation}
such that $\lim_{\kappa \to \infty} h(x) = \min_{i=1,\cdots,I} h_i(x)$. 
The corresponding set $C=\left\lbrace x\in \mathbb{R}^n:h(x)\ge 0\right\rbrace$ lies inside the intersection, $C \subseteq \bigcap_{i=1,\cdots,I} C_i$, such that $\lim_{\kappa \to \infty} C = \bigcap_{i=1,\cdots,I} C_i$.

See Proof of Theorem 4 in \cite{Molnar23CSL}.
$h(x)\ge0$ guarantees $ \min_{i=1,\cdots,I}h_i(x)\ge0$, indicating all constraints $h_i\ge0, i=1,\cdots,I$ are satisfied.

\subsection{Closed-form solution for CBF-based QP}

The safety-oriented framework offers a QP-based optimization approach to modify a nominal policy to ensure safety. 
The nominal policy $\bar{u}$, typically designed to achieve a specific task objective, can be derived from model-based control or generated through RL. 
Based on the established composite CBF $h(x)$ in \eqref{composite}, the optimization problem ensuring system safety can be formulated as the following QP:
\begin{equation} \label{QP2}
	u_s(x) = \underset{{u} \in \mathbb{R}^m}{\arg \min} \frac{1}{2} \|{u} - \bar{u}(x)\|_2^2
\end{equation}
subject to
\begin{equation}\label{QP2cons}
	L_f h(x) + L_g h(x) {u} \geq -\alpha(h(x)).
\end{equation}

When the nominal policy satisfies the safety constraint, the constraint \eqref{QP2cons} is inactive, and the safe policy aligns with the nominal policy. However, when the nominal policy violates the safety constraint, the QP seeks a safe policy that satisfies the constraints while deviating minimally from the nominal policy. 
The purpose of transforming multiple constraints into a composite CBF is to derive a closed-form solution for the safe policy of the optimization \eqref{QP2}. 
The closed-form solution can be obtained by referring to the following theorem.

\textbf{Theorem 1:} Let $C$ be the 0-superlevel set of a continuously differentiable function $h : \mathbb{R}^n \rightarrow \mathbb{R}$, and let $\bar{u}(x) : \mathbb{R}^n \rightarrow \mathbb{R}^m$ be a nominal controller. 
If $h$ is a composite CBF for \eqref{affine} on the set $C \subseteq \bigcap_{i=1,\cdots,I} C_i$ with the corresponding function $\alpha \in \mathcal{K}_{\infty}^e$, then the optimization problem in \eqref{QP2} is feasible for any $x \in \mathbb{R}^n$ and has a closed-form solution given by
\begin{equation}\label{us} % \mathbf{k}_{\text{QP}}(x) 
	u_s(x) = \bar{u}(x)  + \max \{0, \eta(x)\} L_g h(x)^\top
\end{equation}
where the function $\eta : \mathbb{R}^n \rightarrow \mathbb{R}$ is defined as
\begin{equation}\label{etax}
	\eta(x) = \begin{cases} 
		-\frac{L_f h(x) + L_g h(x) \bar{u}(x) + \alpha(h(x))}{\|L_g h(x)\|_2^2} & \text{if } L_g h(x) \neq 0, \\
		0 & \text{if } L_g h(x) = 0.
	\end{cases}
\end{equation}

See proof of Theorem 2 in \cite{Alan23TCST}.
Theorem 2 provides a sufficient but not necessary condition for a safe solution, offering an analytical form for solving the QP associated with a single constraint. 
This formulation eliminates the need to invoke a QP solver, significantly reducing the computational cost. 
Therefore, this advantage motivates its integration with the RL framework. %Moreover, the closed-form solution offers the additional benefit of bypassing the RL framework's requirements for differentiable architectures, 
Furthermore, the closed-form solution provides the significant advantage of circumventing the requirement for differentiable optimizations within the RL framework,
thereby substantially simplifying the gradient computation in the safe policy generation and alleviating the complexity associated with gradient-based optimization, as will be elaborated in the next subsection.
%Furthermore, $\mathbf{k}_{\text{QP}}$ is continuous and $\mathbf{k}_{\text{QP}}(x) \in K_{\text{CBF}}(x)$ for all $x \in \mathbb{R}^n$.

\begin{figure}[h]
	\centering
	\includegraphics[scale=0.47]{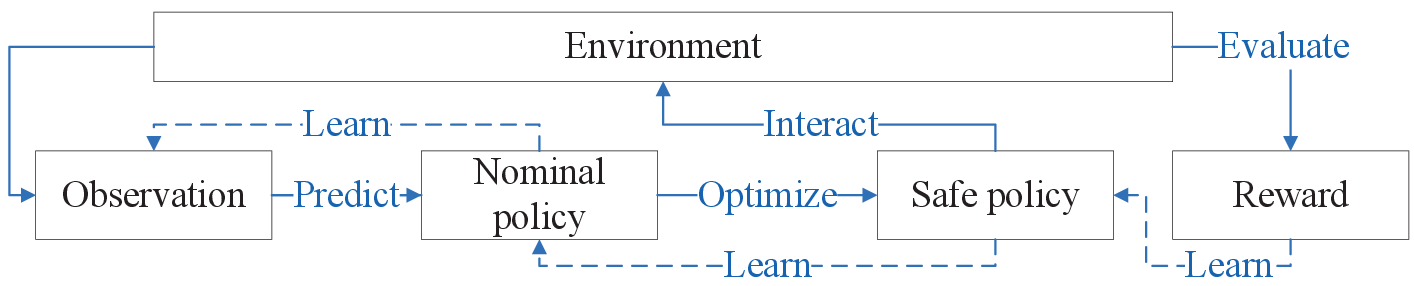}
	\caption{An illustration of an end-to-end training safe RL framework.}
	\label{framework}
\end{figure}

\subsection{Safety layer via closed-form solution in RL framework}
\begin{figure}[h]
	\centering
	\includegraphics[scale=0.34]{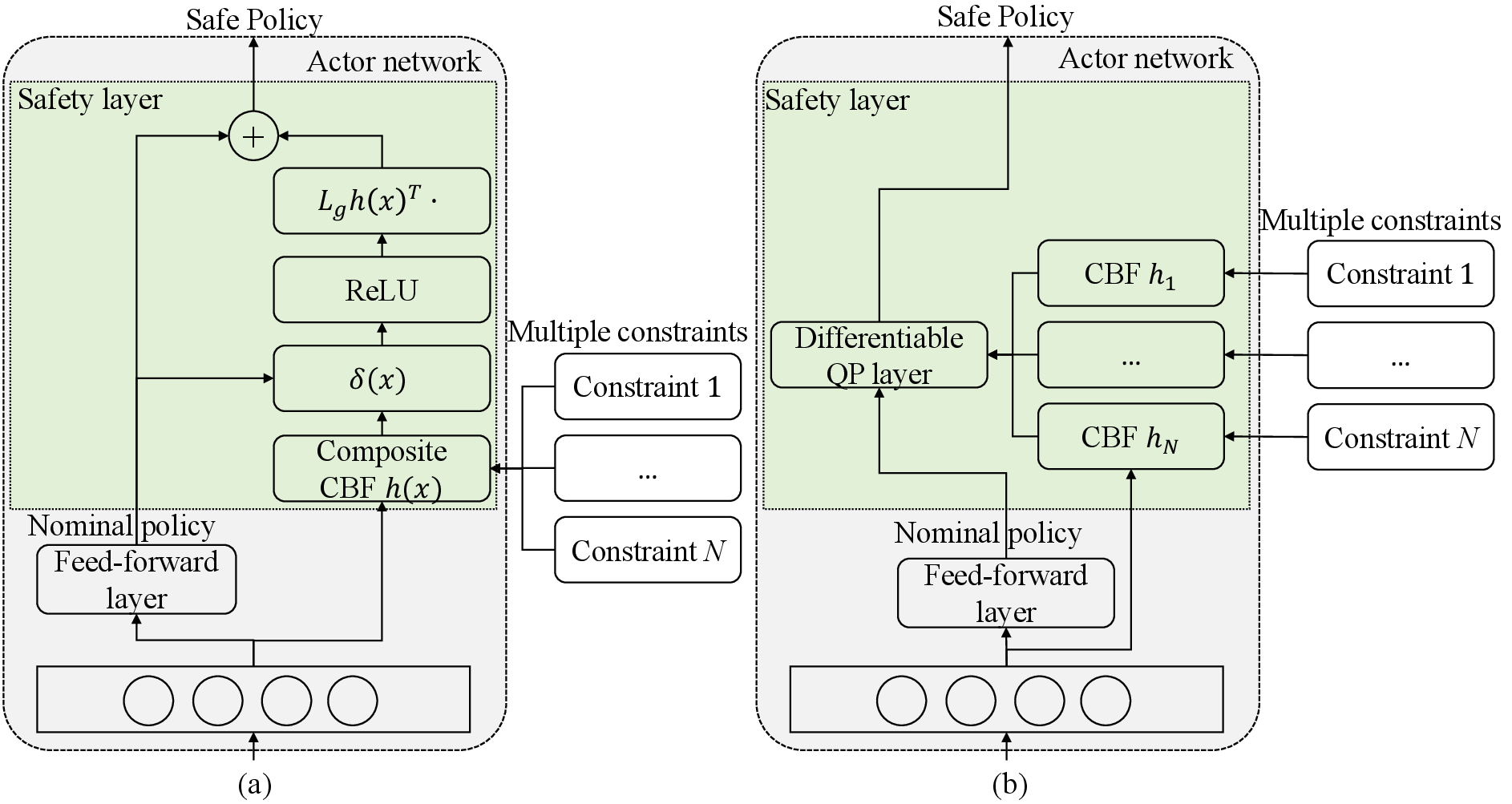}
	\caption{An illustration of safe policy networks with different safety layers. Subfigure (a) demonstrates the proposed framework where $N$ constraints are composited to $h(x)$ using a continuous Log-Sum-Exp approximation. The safety layer is analytical based on the closed-form solution of the composite CBF-based optimization. Subfigure (b) demonstrates the existing framework with differentiable QP layer. The safety layer solves the forward CBF-based optimization and computes the gradient during backpropagation.}
	\label{NNnetwork}
\end{figure}

In conventional RL architectures, the final layer of the control policy network typically consists of a fully connected layer, particularly for continuous control actions in affine systems. 
The output is bounded by the final activation function, such as the hyperbolic tangent, to ensure bounded action outputs.
For safe policy generation, an intuitive approach is to correct the RL-derived control policy by adjusting it through safety-oriented mechanisms, such as correcting the control policy to a safe policy using a CBF-based QP \cite{Cheng_Orosz_Murray_Burdick_2019}. 
However, in this approach, the reward from the safe output cannot backpropagate to the RL network, due to the absence of a gradient pathway connecting the safe policy to the RL policy.
Recently, decision-focused learning have proposed architectures based on differentiable optimization, embedding optimizable and differentiable structures to achieve an end-to-end learning pipeline, thus enabling CBF-QP-based safe learning, as illustrated in Figure \ref{framework}. 
This raises a critical challenge: each training step requires solving a batch of QP problems for the policy loss function, along with calculating the gradient of each QP output concerning the QP parameters, making it computationally expensive and challenging to large-scale multi-constraint problems. 
To address this issue, we integrate a closed-form solution for the safe policy directly into the RL policy generation pipeline. 
This approach leverages a composite single-constraint approximation to handle multi-constraint scenarios, alongside explicit QP solutions to circumvent forward optimization and its gradient backpropagation.
We replace the final layer in RL policy generation with an analytically computed ``safety layer'', which, due to its analytical properties, can be integrated into any actor-critic RL method. 
An illustration of safe policy networks in actor-critic framework is shown in Figure \ref{NNnetwork} with different safety layers.
The proposed framework is demonstrated in 
Figure \ref{NNnetwork}(a), where the closed-form solution \eqref{us} and \eqref{etax} are integrated into the final layer before safety policy generation.
As a comparison, Figure \ref{NNnetwork}(b) demonstrates that taking nominal policy as the input, the differentiable QP layer compute the forward solution with multiple constraints, which is potentially infeasible and computationally expensive.

We illustrate the proposed approach using the SAC method, where the loss functions in this framework are given by 

\begin{equation}
	J_Q(\theta) = \mathbb{E}_{(x_t, u_s^{\phi}) \sim \mathcal{D}_R} \left[ \frac{1}{2} \left( Q_\theta(x_t, u_s^{\phi}) - \left( r(x_t, u_s^{\phi}) + \gamma \mathbb{E}_{x_{t+1} \sim p} \left[ V_{\bar{\theta}}(x_{t+1}) \right] \right) \right)^2 \right],
\end{equation}

\begin{equation}
	V_{\bar{\theta}}(x_t) = \mathbb{E}_{u_s^{\phi} \sim \pi_\phi} \left[ Q_{\bar{\theta}}(x_t, u_s^{\phi}) - \alpha_e \log \pi_\phi (u_s^{\phi} | x_t) \right],
\end{equation}

\begin{equation}\label{Policy_loss_proposed}
	\begin{aligned}
		J_{\pi}(\phi) =  \mathbb{E}_{x_t\thicksim \mathcal{D}_r}  
		\left[ \mathbb{E}_{u_s^{\phi}\thicksim \pi_{\phi}}  [\alpha_e\log \pi_{\phi}(u_s^{\phi}|x_t)-Q_{\theta}(x_t,u_s^{\phi})]               \right],
	\end{aligned}
\end{equation}  %
where $\pi_{\phi}$ denotes the policy generated by the entire policy network, including both the fully connected layers and the QP-based adjustment.
%where $u_t^C$ is the compensation term computed by differentiable QP layer.
In this case, \( u_s^{\phi} \sim \pi_{\phi} \) would mean that the sample \( u_s^{\phi} \) is drawn from the distribution defined by the entire policy network, which inherently includes the safety layer for the QP adjustment. 
%This approach keeps the notation compact and still correctly represents the process, provided it is clear that \(\pi_{\phi}\) now encapsulates both components.

\section{Experiment}
In this section, we aim to validate the capability of the proposed method to ensure safety during training while achieving faster training efficiency compared to the differentiable QP-based approach. The testing environment is designed as a reachability task, where the agent's objective is to reach the goal position while avoiding obstacles. 
To illustrate the incorporation of multiple constraints in the safety-oriented optimization, the collision-free constraints are defined as follows:
\begin{equation}\label{} % h_1(x)=  p_x - p_y > 0, \quad
	\begin{aligned}
		h_i(x) = \|p - p_{i,\text{obs}}\|^2 - r_{\text{safe}}^2 \ge 0, \quad i = 1, \dots, I,
	\end{aligned}
\end{equation}
where $p=[p_x,p_y]^\top$ denotes the position of the agent, $p_{i,\text{obs}}$ denotes the position of the $i$th obstacle, and $r_{\text{safe}}$ is the safe radius to avoid collision.
%The reward function for the SAC process is defined as 
Furthermore, within the safety constraints based on CBFs, the selection of the class-$\mathcal{K}$ function $\alpha$ significantly affects the conservativeness of safety enforcement and $\kappa$ affects the the approximation error of $\min$ operation. 
In this study, we adopt a trade-off value to balance safety and performance with $\alpha=5(\cdot)$, $\kappa=2$.

\begin{figure}[h]
	\centering
	\includegraphics[width=0.7\textwidth]{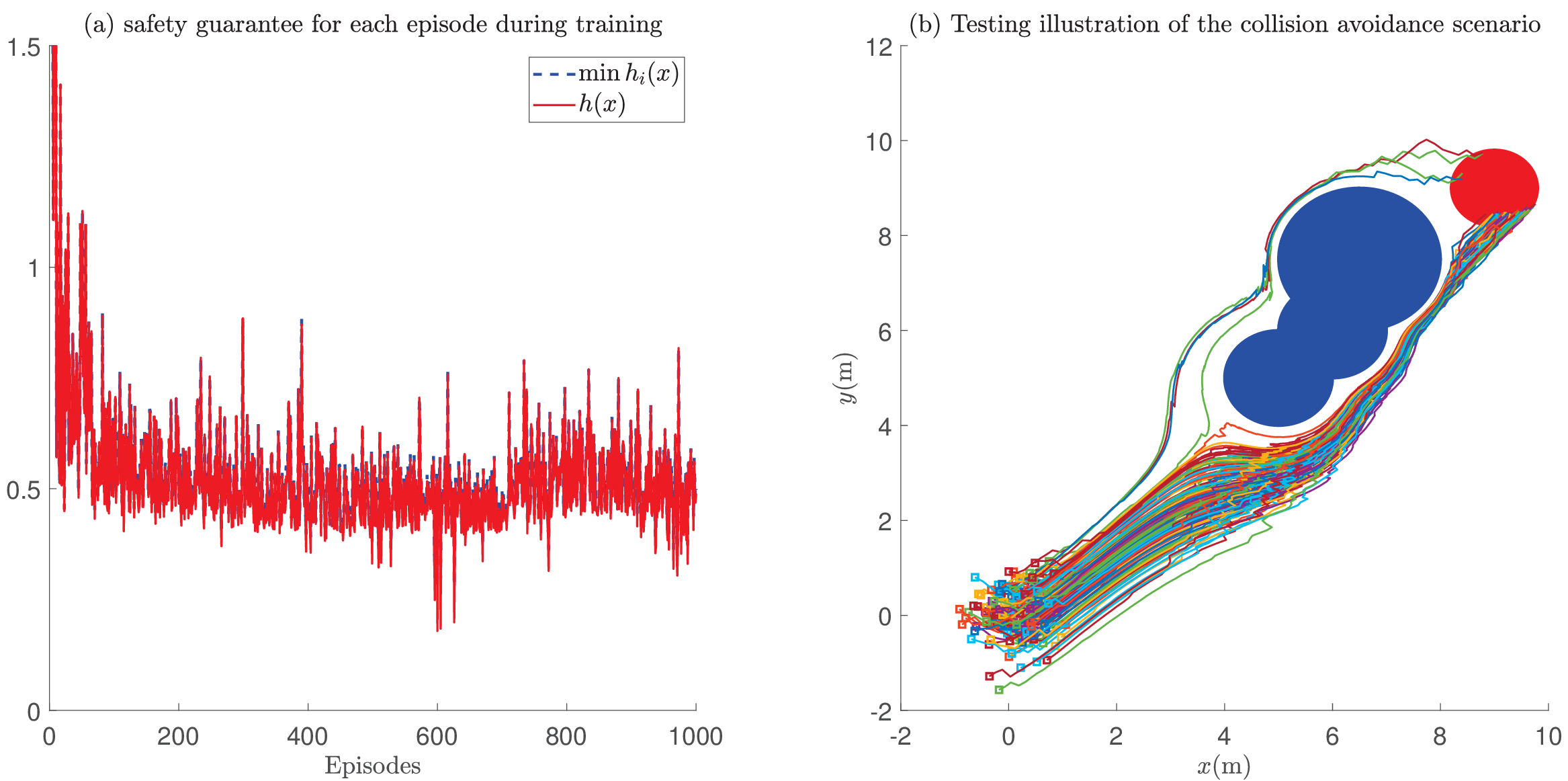}
	\captionsetup{justification=raggedright} % 设置标题左对齐
	\caption{Performance of safe RL training and testing with the proposed method. Subfigure (a) illustrates $\min h_i, i = 1, \dots, I$ and the composite $h$ for each episode during training. The composite $h(x)$ under approximates $\min h_i$ and maintains positive in each episode. Subfigure (b) illustrates the successful trajectories during testing.
		The blue area in Subfigure (b) contains three obstacles, each with a different safe radius size.
		The colored squares denote the initial positions.
		The red circle represents the target area, while the colored lines indicate the testing trajectories, each starting from different initial positions near the origin.} 
	\label{minh_episode}
\end{figure}

\begin{figure}[h]
	\centering
	\includegraphics[width=1.0\textwidth]{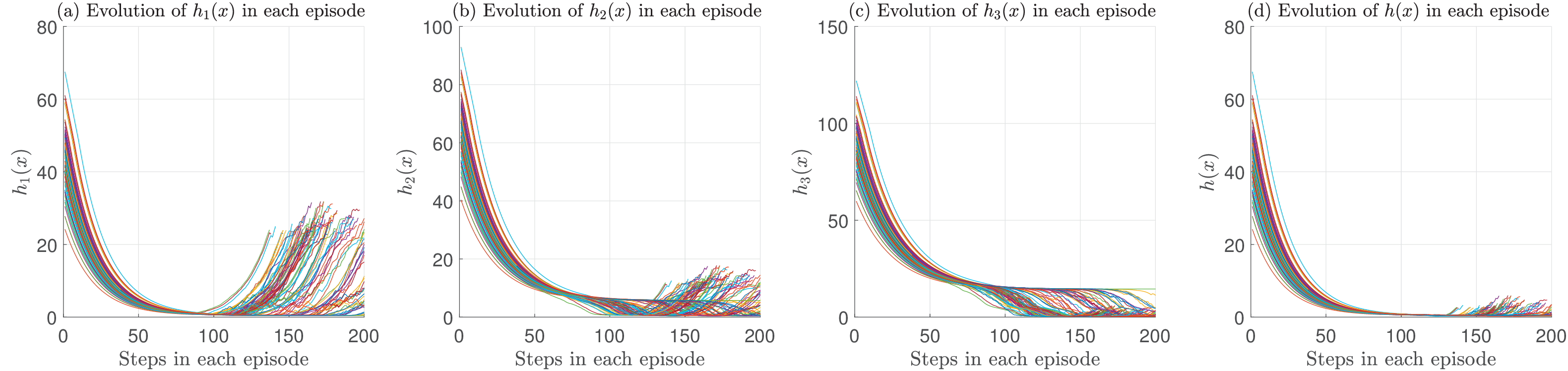}
	\caption{Evolution of $h_1,h_2,h_3$ and the composite $h$ over 1000 episodes. The safe learning during training is guaranteed.}
	\label{minh_steps}
\end{figure}

To demonstrate the safety of the proposed method, we present the following metrics during training: $\min h_i, i = 1, \dots, I$ and the composite $h$ for each episode with $I=3$.
Moreover, the $\min h_i,i\in I$ and composite $h$ across all steps of training episodes, and the trajectories for the deployment phase after training are also demonstrated.
The results are illustrated in the Figure~\ref{minh_episode} and Figure~\ref{minh_steps}.

As shown in Figure~\ref{minh_episode}(a), the minimum value of $h_i$ remains consistently greater than 0 throughout the entire training process, indicating that the system successfully achieves safe training and policy learning. Furthermore, the red line $h(x)$ serves as an under-approximation of the blue dashed line, which is consistent with the conclusion of Theorem 1.
Figure~\ref{minh_episode}(b)  illustrates the trajectories reaching the target area, where safety is ensured across 200 trials. 

A more detailed perspective is provided in Figure~\ref{minh_steps}, which illustrates the evolution of $ h_i $ and $ h $ across all steps in each episode (different color of curves) over $1000$ training iterations (with a maximum step limit of 200). 
The time intervals during which the $ h_i $ curves near the safety set boundary exhibit ``flattening'' correction, depending on the different positions of obstacles. 
During these intervals, the condition $L_f h(x) + L_g h(x) \bar{u}(x) + \alpha(h(x)) < 0 $ holds.
Therefore, the safe policy is actively filtered and corrected based on \eqref{us} and \eqref{etax}, ensuring that $ h $ remains positive for all time.

\begin{table*}[h] % 
	\centering
	\caption{Comparison of different approaches}
	\begin{tabular}{|l|c|c|c|}
		\hline
		\textbf{Method}     & \textbf{ATTS$(s)$ with $I=3$ } & \textbf{ATTS$(s)$ with $I=10$} & \textbf{ATTS$(s)$ with $I=30$} \\ \hline
		\multirow{1}{*}{Closed-form solution}  & 0.018 & 0.024& 0.043 \\  \hline
		\multirow{1}{*}{CBF Batch QP }  & 0.13 & 0.25 & 0.40\\  
		\hline
		\multirow{1}{*}{CBF CVXPYlayer } & 0.84 & 1.45 & 2.26 \\ % CVX fail to work in I=30
		\hline
	\end{tabular}
	\label{table_approach_comparison}
\end{table*}
As previously noted, the closed-form solution eliminates the need for differentiable QP solvers, thereby reducing computational costs. This represents another significant advantage of the proposed method.
In the same scenario for collision avoidance of multiple obstacles, we compared the proposed method with the differentiable QP solver Batched-QPFunction \cite{amos2017optnet} and the CVXPYLayer \cite{agrawal2019differentiable} in terms of computational performance, which are commonly used in similar works within CBF-based safe learning \cite{Emam22RAL,Ma22ECC,jiang2024differentiable,romero2024actor}.
The comparative results are summarized in Table~\ref{table_approach_comparison}, where the performance metric is the average solving time per time step (ATTS) during the RL training process. 
In addition, scenarios with 10 and 30 constraints are also tested to validate the scalability of the proposed method in solving larger-scale safe RL problems.

As demonstrated in Table~\ref{table_approach_comparison}, the proposed method exhibits a computational speed advantage of at least one order of magnitude because of the close-form nature.
The CVXPYlayer-based method, while supporting disciplined parametrized programming, exhibits the lowest solving efficiency due to the lack of support for batch solving and the requirement for gradient computation in QP.
The advantage in training time makes it potentially effective for optimization in large-scale safe RL problems.

\section{Conclusion}

This paper addresses the challenges of ensuring multiple safety constraints and improving training efficiency in safe RL. 
We propose a safe RL framework based on the closed-form solution of composite CBF. 
The framework constructs a composite CBF by using the Log-Sum-Exp approximation of the min function to integrate multiple safety constraints in the optimization problem. 
It also inherits the safety guarantees based on the composite CBF defining the safe set. 
By serving as a surrogate of the differentiable QP architecture with a closed-form solution, the proposed method significantly enhances training efficiency. 
Comparative experiments demonstrate that the proposed method is up to 7 times faster than the current state-of-the-art differentiable batch QP solvers, and at least 46 times faster than the differentiable convex optimization layers CVXPYlayer, showcasing its potential for solving optimization in large-scale safe RL problems.
Future work will further investigate the composite CBF-QP under explicit input constraints, with a focus on guaranteeing feasibility and improving efficiency within the framework of differentiable optimization.

%\acks{We thank a bunch of people.}
\newpage
\acks{This work was supported by National Natural Science Foundation of
	China under Grant 62303316, 
	in part by the Science
	Center Program of National Natural Science Foundation of China under
	Grant 62188101, 
	in part by the Fellowship
	of China National Postdoctoral Program for Innovative Talents under
	Grant BX20240224,
	and the Oceanic Interdisciplinary Program of Shanghai Jiao Tong University (project number
	SL2022MS010).}
	
\bibliography{ref}

\end{document}